\documentclass[11pt,a4paper]{article}
\usepackage[hyperref]{eacl2021}
\usepackage{times}
\usepackage{latexsym}

\usepackage{microtype}

\aclfinalcopy 

\usepackage[T1]{fontenc}
\usepackage{type1cm}
\usepackage[american]{babel}
\usepackage{url}
\usepackage{amssymb}
\usepackage{amsmath}
\usepackage{multirow}
\usepackage{xspace}
\DeclareMathAlphabet{\mathcalbf}{OMS}{pzc}{b}{n}
\usepackage{microtype}
\usepackage[utf8]{inputenc}
\usepackage{enumitem}
\usepackage{scrextend}
\usepackage{graphicx}
\usepackage{csquotes}
\usepackage{hyperref}
\usepackage{scrextend}
\usepackage[multiple]{footmisc}
\usepackage{booktabs}

\RequirePackage{color}
\definecolor{darkgray}{gray}{0.40}
\definecolor{mediumgray}{gray}{0.60}
\definecolor{lightgray}{gray}{0.95}
\definecolor{ultralightgray}{gray}{0.98}

\usepackage{graphicx}
\DeclareGraphicsExtensions{.pdf,.ai,.jpg,.png}
\setkeys{Gin}{pagebox=artbox}
\graphicspath{{./img/}}

\newcommand{\bsfigure}[3][]{%
	\begin{figure}[t]
		\centering
		\includegraphics[#1]{#2}
		\caption{#3}\label{#2}%
 	 \end{figure}
}

\begin{document}
\title{Learning From Revisions: \\ Quality Assessment of Claims in Argumentation at Scale}

\author{Gabriella Skitalinskaya \qquad Jonas Klaff  \\
  Department of Computer Science \\
  University of Bremen \\
  Bremen, Germany \\
  \texttt{\{gabski, joklaff\}@uni-bremen.de} \\
  
  \And
  
  \qquad\qquad{}Henning Wachsmuth \\
  \qquad\qquad{}Department of Computer Science \\
  \qquad\qquad{}Paderborn University \\
  \qquad\qquad{}Paderborn, Germany \\
  \qquad\qquad{}\texttt{henningw@upb.de} \\}
  
\date{}

\maketitle
\urlstyle{rm}

\begin{abstract}
Assessing the quality of arguments and of the claims the arguments are composed of has become a key task in computational argumentation. However, even if different claims share the same stance on the same topic, their assessment depends on the prior perception and weighting of the different aspects of the topic being discussed. This renders it difficult to learn topic-independent quality indicators. In this paper, we study claim quality assessment irrespective of discussed aspects by comparing different {\em revisions of the same claim}. We compile a large-scale corpus with over 377k claim revision pairs of various types from {\em kialo.com}, covering diverse topics from politics, ethics, entertainment, and others. We then propose two tasks: (a)~assessing which claim of a revision pair is better, and (b)~ranking all versions of a claim by quality. Our first experiments with embedding-based logistic regression and transformer-based neural networks show promising results, suggesting that learned indicators generalize well across topics. In a detailed error analysis, we give insights into what quality dimensions of claims can be assessed reliably. We provide the data and scripts needed to reproduce all results.%
\footnote{Data and code:\url{https://github.com/GabriellaSky/claimrev}}
\end{abstract}

\section{Introduction}
\label{sec:introduction}

Assessing argument quality is as important as it is questionable in nature. On the one hand, identifying the good and the bad claims and reasons for arguing on a given topic is key to convincingly support or attack a stance in debating technologies \cite{rinott:2015}, argument search \cite{ajjour:2019}, and similar. On the other hand, argument quality can be considered on different granularity levels and from diverse perspectives, many of which are inherently subjective \cite{wachsmuth-etal-2017-computational}; they depend on the prior beliefs and stance on a topic as well as on the personal weighting of different aspects of the topic \cite{kock:2007}.

\begin{table*}[t]
\small
\centering
\setlength{\tabcolsep}{5pt}
\begin{tabular}{p{5.3cm} p{7cm} p{2.3cm}}
\toprule
\bf Claim before Revision & \bf Claim after Revision & \bf Type \\ 
\midrule
Dogs  can  help disabled  people  function better. & Dogs can help disabled people to navigate the world better. & Claim Clarification \\
\addlinespace
 African American soldiers joined unionists to fight for their freedom. & Black soldiers joined unionists to fight for their freedom. & Typo / Grammar Correction\\ 
\addlinespace
Elections insure the independence of the judiciary.		& Elections ensure the independence of the judiciary.	& Typo / Grammar Correction \\
\addlinespace
Israel has a track record of selling US arms to third countries without authorization. & Israel has a track record of selling US arms to third countries without authorization ({\em \scriptsize https://www.jstor.org/stable/1149008?seq=1\#page\_scan\_tab\_contents}). & Corrected / Added links \\
\bottomrule
\end{tabular}
\caption{Four examples of claims from Kialo before and after revision, along with the type of revision performed.}
\label{tab:pair_example}
\end{table*}

Existing research largely ignores this limitation, by focusing on learning to predict argument quality based on subjective assessments of human annotators (see Section~\ref{sec:relatedwork} for examples). In contrast, \newcite{habernal-gurevych-2016-argument} control for topic and stance to compare the convincingness of arguments. \newcite{wachsmuth-etal-2017-pagerank} abstract from an argument's text, assessing its relevance only structurally.  \newcite{lukin:2017} and \newcite{elbaff:2020} focus on personality-specific and ideology-specific quality perception, respectively, whereas \newcite{toledo:2019} asked annotators to disregard their own stance in judging length-restricted arguments. However, none of these approaches controls for the concrete aspects of a topic that the arguments claim and reason about. This renders it difficult to learn what makes an argument and its building blocks good or bad in general.

In this paper, we study quality in argumentation irrespective of the discussed topics, aspects, and stances by assessing different revisions of the basic building blocks of arguments, i.e., claims. Such revisions are found in large quantities on online debate platforms such as {\em kialo.com}, where users post claims, other users suggest revisions to improve claim quality (in terms of clarity, grammaticality, grounding, etc.), and moderators approve or disapprove them. By comparing the quality of different revisions of the same instance, we argue that we can learn general quality characteristics of argumentative text and, to a wide extent, abstract from prior perceptions and weightings.

To address the proposed problem, we present a new large-scale corpus, consisting of 124k unique claims from kialo.com spanning a diverse range of topics related to politics, ethics, and several others (Section~\ref{sec:data}). Using distant supervision, we derive a total number of 377k claim revision pairs from the platform, each reflecting a quality improvement, often, with a specified revision type. Four examples are shown in Table \ref{tab:pair_example}. To the best of our knowledge, this is the first corpus to target quality assessment based on claim revisions. In a manual annotation study, we provide support for our underlying hypothesis that a revision improves a claim in most cases, and we test how much the revision types correlate with known argument quality dimensions. 

Given the corpus, we study two tasks: (a)~how to compare revisions of a claim by quality and (b)~how to rank a set of claim revisions. As initial approaches to the first task, we select in Section~\ref{sec:approach} a ``traditional'' logistic regression model based on word embeddings as well as transformer-based neural networks \cite{Vaswani2017Attention}, such as BERT \cite{devlin-etal-2019-bert} and SBERT \cite{reimers-gurevych-2019-sentence}. For the ranking task, we consider the Bradley-Terry-Luce model \cite{bradley_1952_BT, luce2012individual} and SVMRank \cite{joachims_svm2006}. They achieve promising results, indicating that the compiled corpus allows learning topic-independent characteristics associated with the quality of claims (Section~\ref{sec:experiments}). To understand what claim quality improvements can be assessed reliably, we then carry out a detailed error analysis for different revision types and numbers of revisions.

The main contributions of our work are: (1)~A new corpus for topic-independent claim quality assessment, with distantly supervised quality improvement labels of claim revision pairs, (2)~initial promising approaches to the tasks of claim quality classification and ranking, and (3)~insights into what works well in claim quality assessment and what remains to be solved.

\section{Related Work}
\label{sec:relatedwork}

In the recent years, there has been an increase of research on the quality of arguments and the claims and reasoning they are composed of. \newcite{wachsmuth-etal-2017-computational} describe argumentation quality as a multidimensional concept that can be considered from a logical, rhetorical, and dialectical perspectives. To achieve a common understanding, the authors suggest a unified framework with 15 quality dimensions, which together give a holistic quality evaluation at a certain abstraction level. They point out, that several dimensions may be perceived differently depending on the target audience. In recent follow-up work, \newcite{wachsmuth:2020} examined how well each dimension can be assessed only based on plain text only.

Most existing quality assessment approaches target a single dimension. On mixed-topic student essays, \newcite{persing-ng-2013-modeling} learn to score the clarity of an argument's thesis, \newcite{persing-ng-2015-modeling} do the same for argument strength, and \newcite{stab:2017} classify whether an argument's premises sufficiently support its conclusion. All these are trained on pointwise quality annotations in the form of scores or binary judgments. \citet{gretz2019large} provide a corpus with crowdsourced quality annotations for 30,497 arguments, the largest to date for pointwise argument quality. The authors studied how their annotations correlate with the 15 dimensions from the framework of \citet{wachsmuth-etal-2017-computational}, finding that only {\em global relevance} and \textit{effectiveness} are captured. Similarly, \newcite{lauscher:2020} built a new corpus based on the framework to then exploit interactions between the dimensions in a neural approach. We present a small related annotation study for our dataset below. However, we follow \newcite{habernal-gurevych-2016-argument} in that we cast argument quality assessment as a relation classification problem, where the goal is to identify the better among a pair of instances. 

In particular, \newcite{habernal-gurevych-2016-argument} created a dataset with argument convincingness pairs on 32 topics. To mitigate annotator bias, the arguments in a pair always have the same stance on the same topic. The more convincing argument is then predicted using a feature-rich SVM and a simple bidirectional LSTM. Other approaches to the same task map passage representations to real-valued scores using Gaussian Process Preference Learning \cite{simpson-gurevych-2018-finding} or represent arguments by the sum of their token embeddings \cite{potash-etal-2017-length}, later extended by a Feed Forward Neural Network \cite{potash-etal-2019-ranking}. Recently, \newcite{gleize-etal-2019-convinced} employed a Siamese neural network to rank arguments by the convincingness of evidence. In our experiments below, we take on some of these ideas, but also explore the impact of transformer-based methods such as BERT \cite{devlin-etal-2019-bert}, which have been shown to predict argument quality well \cite{gretz2019large}.

\newcite{potash-etal-2017-length} observed that longer arguments tend to be judged better in existing corpora, a phenomenon we will also check for below. \citet{toledo-etal-2019-automatic} prevent such bias in their corpora for both pointwise and pairwise quality, by restricting the length of arguments to 8--36 words. The authors define quality as the level of preference for an argument over other arguments with the same stance, asking annotators to disregard their own stance. For a more objective assessment of argument relevance, \newcite{wachsmuth-etal-2017-pagerank} abstract from content, ranking arguments only based on structural relations, but they employ majority human assessments for evaluation. \newcite{lukin:2017} take a different approach, including knowledge about the personality of the reader into the assessment, and \newcite{elbaff:2020} study the impact of argumentative texts on people depending on their political ideology.

As can be seen, several approaches aim to control for length, stance, audience, or similar. However, all of them still compare argumentative texts with different content and meaning in terms of the aspects of topics being discussed. In this work, we assess quality based on different revisions of the same text. In this setting, the quality is primarily focused on how a text is formulated, which will help to better understand what influences argument quality in general, irrespective of the topic. To be able to do so, we refer to online debate portals. 

Debate portals give users the opportunity to discuss their views on a wide range of topics. Existing research has used the rich argumentative content and structure of different portals for argument mining, including {\em createdebate.com} \cite{habernal:2015}, {\em idebate.org} \cite{alkhatib:2016}, and others. Also, large-scale debate portal datasets form the basis of applications such as argument search engines \cite{ajjour:2019}. Unlike these works, we exploit debate portals for studying {\em quality}. \newcite{tan2016persuasion} predicted argument persuasiveness in the discussion forum {\em ChangeMyView} from ground-truth labels given by opinion posters, and \newcite{wei-etal-2016-post} used user upvotes and downvotes for the same purpose. Here, we resort to {\em kialo.com}, where users cannot only state argumentative claims and vote on the impact of claims submitted by others, but they can also help improve claims by suggesting revisions, which are approved or disapproved by moderators. While \newcite{durmus-etal-2019-role} assessed quality based on the impact value of claims from kialo.com, we derive information on quality from the revision history of claims.

The only work we are aware of that analyzes revision quality of argumentative texts is the study of \newcite{afrin-litman-2018-annotation}. From the corpus of \newcite{zhang:2017} containing 60 student essays with three draft versions each, 940 sentence writing revision pairs were annotated for whether the revision improves essay quality or not. The authors then trained a random forest classifier for automatic revision quality classification. In contrast, instead of sentences, we shift our focus to claims. Moreover, our dataset is orders of magnitude larger and includes notably longer revision chains, which enables deeper analyses and more reliable prediction of revision quality using data-intensive methods. 
\section{Data}
\label{sec:data}

Here, we present our corpus created based on claim revision histories collected from {\em kialo.com}. 

\subsection{A New Corpus based on Kialo}

Kialo is a typical example of an online debate portal for collaborative argumentative discussions, where participants jointly develop complex pro/con debates on a variety of topics. 
The scope ranges from general topics (religion, fair trade, etc.) to very specific ones, for instance, on particular policy-making (e.g., whether wealthy countries should provide citizens with a universal basic income). Each debate consists of a set of claims and is associated with a list of related pre-defined generic categories, such as politics, ethics, education, and entertainment.

What differentiates Kialo from other portals is that it allows editing claims and tracking changes made in a discussion. All users can help improve existing claims by suggesting edits, which are then accepted or rejected by the moderator team of the debate. As every suggested change is discussed by the community, this collaborative process should lead to a continuous improvement of claim quality and a diverse set of claims for each topic.

As a result of the editing process, claims in a debate have a version history in the format of claim pairs, forming a chain where one claim is the successor of another and is considered to be of higher quality (examples found in Table \ref{tab:pair_example}). In addition, claim pairs may have a revision type label assigned to them via a non-mandatory free form text field, where moderators explain the reason of revision.

\paragraph{Base Corpus}

To compile the corpus, we scraped all 1628 debates found on Kialo until June 26th, 2020, related to over 1120~categories. They contain 124,312~unique claims along with their revision histories, which comprise of 210,222~pairwise relations. The average number of revisions per claim is~1.7 and the maximum length of a revision chain is~36. 74\% of all pairs have a revision type. Overall, there are 8105 unique revision type labels in the corpus. 92\% of labeled claim pairs refer to three types only: {\em Claim Clarification}, {\em Typo/Grammar Correction}, and {\em Corrected/Added Links}. An overview of the distribution of revision labels is given in Table~\ref{tab:dataset_desc}. We refer to the resulting corpus as {\em ClaimRev\textsubscript{\tiny{BASE}}}.

\begin{table}[t]
\small
\centering
\renewcommand{\arraystretch}{1}
\setlength{\tabcolsep}{5pt}
\begin{tabular}{llr}
\toprule
\bf Corpus						& \bf Type of Instances	& \bf Instances	\\
\midrule
ClaimRev\textsubscript{{\tiny{BASE}}}	& \bf Total claim pairs	& \bf 210\,222	\\
 & Claim Clarification	& 63 729	\\
 &   Typo/Grammar Correction	& 59 690		\\
  &  Corrected/Added Links    & 17 882		\\
  &  Changed Meaning of Claim    & 1 178		\\
  &  Misc	& 10 464		\\
  &  None	& 57 279		\\
\midrule
ClaimRev\textsubscript{{\tiny{EXT}}}		& \bf Total claim pairs	& \bf 377\,659	\\
 & Revision distance 1				& 77 217\\
 & Revision distance 2				& 27 819\\
 & Revision distance 3			& 10 753\\
 & Revision distance 4				& 4 460	 \\
 & Revision distance 5				& 2 055 \\
 & Revision distance 6+				& 2 008 \\
 \midrule	
Both Corpora					& \bf Claim revision chains			& \bf 124\,312	\\	
\bottomrule
\end{tabular}
\caption{Statistics of the two provided corpus versions. ClaimRev\textsubscript{{\tiny{BASE}}}: Number of claim pairs in total and of each revision type. ClaimRev\textsubscript{{\tiny{EXT}}}: Number of claim pairs in total and of each revision distance. The bottom line shows the number of unique revision chains in the corpora.}
\label{tab:dataset_desc}
\end{table}

Data pre-processing included removing all claim pairs from debates carried out in languages other than English. Also, we considered claims with less than four characters as uninformative and left them out. As we seek to compare different versions of the {\em same} claim, claim version pairs with a general change of meaning do not satisfy this description. Thus, we removed such pairs from the corpus, too (inspecting the data revealed that such pairs were mostly generated due to debate restructuring).
For this, we assessed the cosine similarity of a given claim pair using {\em spacy.io} and remove a pair if the score is lower than the threshold of 0.8.

\paragraph{Extended Corpus}

\begin{figure}[t]
    \centering
    \includegraphics[width=7.7cm]{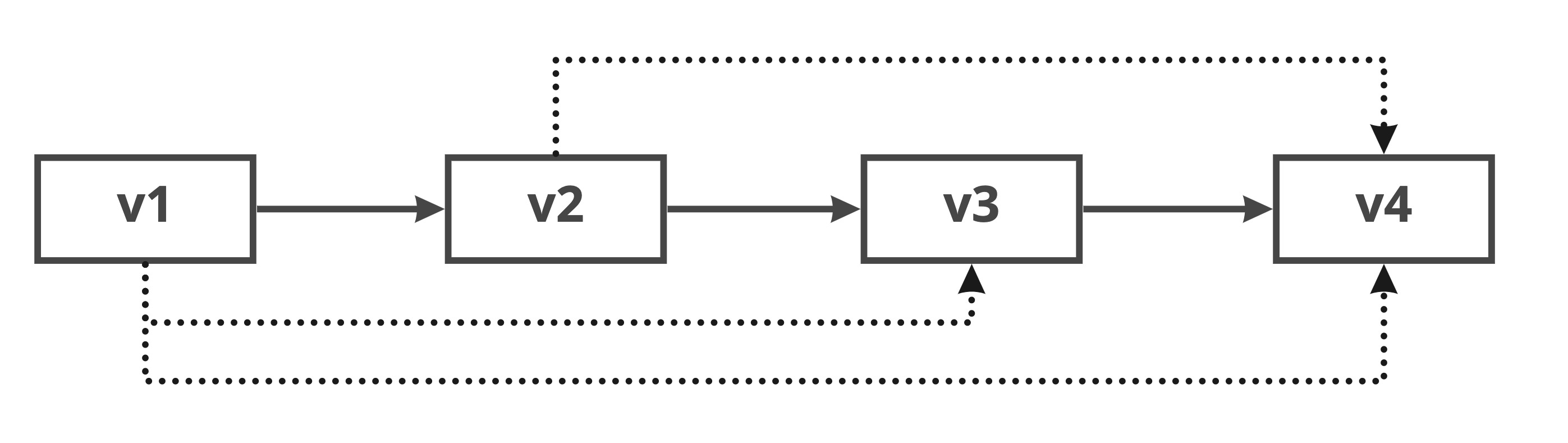}
    \caption{Visual representation of relations between revisions. Solid and dashed lines denote original and inferred non-consecutive relations respectively.}
    \label{fig:relations}
\end{figure}

To increase the diversity of data available for training models, without actually collecting new data, we applied data augmentation. ClaimRev\textsubscript{{\tiny{BASE}}} consists of consecutive claim version pairs, i.e., if a claim $v$ has four versions, it will be represented by three three pairs: $(v_1, v_2)$, $(v_2, v_3)$, and $(v_3, v_4)$, where $v_1$ is the original claim and $v_4$ is the latest version. We extend this data by adding all pairs between non-consecutive versions that are inferrable transitively. Considering the previous example, this means we add $(v_1,v_3)$, $(v_1, v_4)$, and $(v2,v4)$. This is based on our hypothesis that every argument version is of higher quality than its predecessors, which we come back to below. Figure~\ref{fig:relations} illustrates the data augmentation. We call the augmented corpus {\em ClaimRev\textsubscript{{\tiny{EXT}}}}. 

For this corpus, we introduce the concept of {\em revision distance}, by which we mean the number of revisions between two versions. For example, the distance between $v_1$ and $v_2$ would be 1, whereas the distance between $v_1$ and $v_3$ would be 2. The distribution of the revision distances across ClaimRev\textsubscript{{\tiny {EXT}}} is summarized in Table~\ref{tab:dataset_desc}.

\medskip
The number of claim pairs of the 20 most frequent categories in both corpus versions are presented in Figure \ref{topic-distribution}. We will restrict our view to the topics in these categories in our experiments.

\bsfigure{topic-distribution}{Number of claim revision pairs in each debate category of the two provided versions of our corpus  (ClaimRev\textsubscript{{\tiny{BASE}}}, ClaimRev\textsubscript{{\tiny{EXT}}}).}

\subsection{Data Consistency on Kialo}
\label{sec:data-consistency}

While collaborative content creation enables leveraging the wisdom of large groups of individuals toward solving problems, it also poses challenges in terms of quality control, because it relies on varying perceptions of quality, backgrounds, expertise, and personal objectives of the moderators. To assess the consistency of the distantly-supervised corpus annotations, we carried out two annotation studies on samples of our corpus.

\paragraph{Consistency of Relative Quality}

In this study, we aimed to capture the general perception of claim quality on a meta-level, by deriving a data-driven quality assessment based on the revision histories. This was based on our hypothesis that every claim version is better than its predecessor. To test the validity of this hypothesis, two authors of this paper annotated whether a revision increases, decreases, or does not affect the overall claim quality. For this purpose, we randomly sampled 315 claim revision pairs, found in the supplementary material.

The results clearly support our hypothesis, showing an increase in quality in 292 (93\%) of the annotated cases at a Cohen's $\kappa$ agreement of 0.75, while 8 (3\%) of the revisions had no effect on quality and only 6 (2\%) led to a decrease. On the remaining 2\%, the annotators did not reach an agreement.

\paragraph{Consistency of Revision Type Labels }

Our second annotation study focused on the reliability of the revision type labels. We restricted our view to the top three revision labels, which cover 96\% of all revisions.
We randomly sampled 140--150 claim pairs per each revision type, 440 in total. For each claim pair, the same annotators as above provided a label for the revision type from the following set: {\em Claim Clarification}, {\em Typo/Grammar Correction}, {\em Corrected/Added Links}, and {\em Other}.

Comparing the results to the original labels in the corpus revealed that the annotators strongly agreed with the labels, namely, with Cohen's $\kappa$ of 0.82 and 0.76 respectively. The level of agreement between the annotators was even higher ($\kappa$ = 0.84). In further analysis, we observed that most confusion happened between the revision types {\em Typo/Grammar correction} and {\em Claim Clarification}. This may be due to the non-strict nature of the revision type labels, which leaves space for different interpretations on a case-to-case basis. Still, we conclude that the revision type labels seem reliable in general.

\subsection{Quality Dimensions on Kialo}

To explore the relationship between the revision types on Kialo and argument quality in general, we conducted a third annotation study. In particular, for each of the 315 claim pairs from Section \ref{sec:data-consistency}, one of the authors of this paper provided a label indicating whether the revision improved for each of the 15 quality dimensions defined by \newcite{wachsmuth-etal-2017-computational} or not. It should be noted that the annotators reached an agreement on the revision type for all these pairs.

\begin{table}[t]
\centering
\small
\setlength{\tabcolsep}{3pt}
\renewcommand{\arraystretch}{0.945}
\begin{tabular}{l@{$\!\!\!\!\!\!\!\!\!\!\!\!$}rr@{\quad\quad\,\,}r}
\toprule
 & \textbf{Clarification} & \textbf{Grammar} & \textbf{Links} \\
 \midrule
 \bf Cogency & \textbf{-0.31} & \textbf{-0.31} & \textbf{0.65} \\
Local Acceptability & \textbf{0.38} & -0.20 & -0.19 \\
Local Relevance & \textbf{0.44} & -0.25 & -0.22 \\
Local Sufficiency & -0.28 & \textbf{-0.33} &\bf 0.62 \\
\addlinespace
\bf Effectiveness & 0.02 & \textbf{-0.35} & \textbf{0.34} \\
Credibility & 0.06 & -0.16 & 0.10 \\
Emotional Appeal & 0.00 & 0.00 & 0.00 \\
Clarity & -0.16 & \textbf{0.35} & -0.18 \\
Appropriateness & 0.01 & 0.02 & -0.04 \\
Arrangement & 0.00 & 0.00 & 0.00 \\
\addlinespace
\bf Reasonableness & 0.07 & -0.04 & -0.04 \\
Global Acceptability & \textbf{0.37} & \textbf{0.42} & \textbf{-0.82} \\
Global Relevance & 0.02 & \textbf{-0.43} & \textbf{0.42} \\
Global Sufficiency & 0.00 & 0.00 & 0.00 \\
\addlinespace
\bf Overall & -0.05 & 0.00 & 0.05\\
\midrule
Pairs with revision type & 120 &100 & 95\\
\bottomrule
\end{tabular}
\caption{Pearson's $r$ correlation in our annotation study between increases in the 15 quality dimensions of \newcite{wachsmuth-etal-2017-computational} and the main revision types: Claim {\em Clarification}, Typo/{\em Grammar} Correction, Corrected/Added {\em Links}. Moderate and high correlations are shown in bold ($r \geq 0.3$).}
\label{tab:dimension-analysis}
\end{table}

Table \ref{tab:dimension-analysis} shows Pearson's $r$ rank correlation for each quality dimension for the three main revision types. We observe a strong correlation between the revision type {\em Corrected/Added Links} and the logical quality dimensions {\em Cogency} (0.65) and {\em Local Sufficiency} (0.62), which matches the main purpose of such revisions: to add supporting information to a claim. The high negative correlation of this revision type with {\em Global Acceptability} (-0.82) indicates that improvements regarding the dimension in question are more prominent in other types. Complementarily, {\em Claim Clarification} mainly improves the other logical dimensions ({\em Local Acceptability} 0.38, {\em Local Relevance} 0.44), matching the intuition that a clarification helps to ensure a correct understanding of the meaning. {\em Typo/Grammar corrections}, finally, rather seem to support an acceptable linguistic shape, improving {\em Clarity} (0.35) and {\em Global Acceptability} (0.42). 

Finding only low correlations for many rhetorical dimensions (credibility, emotional appeal, etc.) as well as for overall quality, we conclude that the revisions on Kialo seem to target primarily the general form a well-phrased claim should have.
\section{Approaches}
\label{sec:approach}

To study the two proposed tasks, claim quality classification and claim quality ranking, on the given corpus, we consider the following approaches.

\subsection{Claim Quality Classification}

We cast this task as a pairwise classification task, where the objective is to compare two versions of the same claim and determine which one is better. To solve this task, we compare four methods:

\paragraph{Length} 
To check whether there is a bias towards longer claims in the data, we use a trivial method which assumes that claims with more characters are better.

\paragraph{S-BOW} 

As a ``traditional'' method, we employ the siamese bag-of-words embedding (S-BOW) as described by \citet{potash-etal-2017-length}. We concatenate two bag-of-words matrices, each representing a claim version from a pair, and input the concatenated matrix to a logistic regression. We also test whether information on length improves S-BOW.

\paragraph{BERT} 

We select the BERT model, as it has become the standard neural baseline. BERT is a pre-trained deep bidirectional transformer language model \cite{devlin-etal-2019-bert}. 
For our experiments we use the pre-trained version {\em bert-base-cased}, as implemented in the {\em huggingface} library.%
\footnote{Huggingface library, \url{https://huggingface.co/transformers/pretrained_models.html}\label{source}} 
We fine-tune the model for two epochs using the Adam optimizer with learning rate 1e-5.
\footnote{\label{footnote:window}We chose the number of epochs empirically, picking the best learning rate out of \{5e-7, 5e-6,1e-5,2e-5,3e-5\}.}

\paragraph{SBERT} 

 We also use Sentence-BERT (SBERT) to learn to represent each claim version as a sentence embedding \cite{reimers-gurevych-2019-sentence}, opposed to the token-level embeddings of standard BERT models. We fine-tune SBERT based on {\em bert-base-cased} using a siamese network structure, as implemented in the {\em sentence-transformers} library.%
\footnote{Sentence-transformers library, \url{https://www.sbert.net/}\label{source}} 
We set the numbers of epochs to one which is recommended by the authors \cite{reimers-gurevych-2019-sentence}, and we use a batch-size of 16, Adam optimizer with learning rate 1e-5, and a linear learning rate warm-up over 10\% of the training data. Our default pooling strategy is MEAN.

\begin{table}[t]
\small
\centering
\renewcommand{\arraystretch}{1}
\begin{tabular}{c|ccc}
\toprule
 & $v_1$ & $v_2$ &$v_3$ \\ 
\midrule
$v_1$ & 0 & 0.018 & 0.002 \\ 
$v_2$ & 0.982 & 0 & 0.428 \\ 
$v_3$ & 0.998 & 0.572 & 0 \\ 
\bottomrule
\end{tabular}
\caption{Example of a pairwise score matrix for ranking of three claim revisions, $v_1$--$v_3$, given the following pairwise scores: $(v_1,v_2) = (0.018, 0.982)$, $(v_2,v_3) = (0.428, 0.572)$, and $(v_1, v_3) = (0.002, 0.998)$.}
\label{tab:relations}
\end{table}

\subsection{Claim Quality Ranking}

In contrast to the previous task, we cast this problem as a sequence-pair regression task. After obtaining all pairwise scores using S-BOW, BERT, and SBERT respectively, we map the pairwise labels to real-valued scores and rank them using the following models, once for each method. 

\paragraph{BTL} 

\begin{table*}[t]
\small
\centering
\setlength{\tabcolsep}{4.1pt}
\renewcommand{\arraystretch}{1}
\begin{tabular}{l@{$\!$}crrrrcrrrr}
\toprule
			& 	& \multicolumn{4}{c}{\bf Test set: ClaimRev\textsubscript{{\tiny{BASE}}}}						&	& \multicolumn{4}{c}{\bf Test set: ClaimRev\textsubscript{{\tiny{EXT}}}}						\\
			\cmidrule(l@{2pt}r@{2pt}){3-6}											\cmidrule(l@{2pt}r@{2pt}){8-11}
			&	& \multicolumn{2}{c}{\bf Random-Split}	& \multicolumn{2}{c}{\bf Cross-Category}	&	& \multicolumn{2}{c}{\bf Random-Split}	& \multicolumn{2}{c}{\bf Cross-Category}	\\	
			\cmidrule(l@{2pt}r@{2pt}){3-4}				\cmidrule(l@{2pt}r@{2pt}){5-6}				\cmidrule(l@{2pt}r@{2pt}){8-9}			\cmidrule(l@{2pt}r@{2pt}){10-11}		
\bf Model	&	& \bf Accuracy	& \bf MCC				& \bf Accuracy	& \bf MCC				&	& \bf Accuracy	& \bf MCC				& \bf Accuracy	& \bf MCC				\\ 
\midrule
Length		&	& 61.3\,/\,61.3	& 0.23\,/\,0.23	& 60.7\,/\,60.7	& 0.21\,/\,0.21	&	& 60.8\,/\,60.8	& 0.22\,/\,0.22	& 60.0\,/\,60.0	& 0.20\,/\,0.20 \\
SBOW		&	& 62.0\,/\,62.6 & 0.24\,/\,0.25 & 61.4\,/\,61.4 & 0.23\,/\,0.23 & 	& 64.9\,/\,65.4	& 0.30\,/\,0.31 & 63.9\,/\,64.1 & 0.28\,/\,0.28 \\
SBOW + Length&	& 65.1\,/\,65.5 & 0.30\,/\,0.31 & 64.8\,/\,64.4 & 0.29\,/\,0.29 & 	& 67.1\,/\,67.5 & 0.34\,/\,0.35	& 66.1\,/\,66.2 & 0.32\,/\,0.32 \\
BERT		&	& 75.5\,/\,75.2	& 0.51\,/\,0.51	& 75.1\,/\,74.1	& \textbf{0.51}\,/\,0.49	& 	&  76.4\,/\,76.5		&  0.53\,/\,0.53				& 76.2\,/\,75.4 	& 0.53\,/\,0.51 	\\
SBERT		&	& \textbf{76.2}\,/\,\textbf{76.2} & \textbf{0.53}\,/\,\textbf{0.52} &  \bf 75.5\,/\,75.4 & \bf 0.51\,/\,0.51	& 	&  \bf 77.4\,/\,77.7 & 	\bf 0.55\,/\,0.55 & \textbf{76.8}\,/\,\textbf{76.8} & \textbf{0.54}\,/\,\textbf{0.54}	\\
\midrule
Random baseline&	& 50.0\,/\,50.0	& 0.00\,/\,0.00	& 50.0\,/\,50.0	& 0.00\,/\,0.00	&	& 50.0\,/\,50.0	& 0.00\,/\,0.00	& 50.0\,/\,50.0	& 0.00\,/\,0.00	\\
Single claim baseline&	&  57.7\,/\,58.1		& 	0.17\,/\,0.17			& 	57.7\,/\,57.3	&  	0.17\,/\,0.16			& 	& 58.8\,/\,59.8	& 0.20\,/\,0.20	& 	58.9\,/\,58.9		& 	0.20\,/\,0.20	\\
\bottomrule
\end{tabular}
\caption{Claim quality classification results: Accuracy and Matthew Correlation Coefficient (MCC) for all tested approaches in the random-split and the cross-category setting on the two corpus versions. The first value in each value pair is obtained by a model trained on ClaimRev\textsubscript{{\tiny{BASE}}}, the second by a model trained on ClaimRev\textsubscript{{\tiny{EXT}}}. All improvements from one row to the next are significant at $p <$ 0.001 according to a two-sided Student's $t$-test.}

\label{tab:res_class}
\end{table*}

For mapping, we use the well-established Bradley-Terry-Luce (BTL) model \cite{bradley_1952_BT,luce2012individual}, in which items are ranked according to the probability that a given item beats an item chosen randomly. We feed the BTL model a pairwise-comparison matrix for all revisions related to a claim, generated as follows: Each row represents the probability of the revision being better than other revisions. All diagonal values are set to zero. Table \ref{tab:relations} illustrates an example for a set of three argument revisions.

\paragraph{SVMRank} 

Additionally, we employ SVMRank \cite{joachims_svm2006}, which views the ranking problem as a pairwise classification task. First, we change the input data, provided as a ranked list, into a set of ordered pairs, where the (binary) class label for every pair is the order in which the elements of the pair should be ranked. Then, SVMRank learns by minimizing the error of the order relation when comparing all possible combinations of candidate pairs. Given the nature of the algorithm we cannot work with token embeddings obtained from BERT directly. Thus, we utilize one of most commonly used approaches to transform token embeddings to a sentence embedding: extracting the special [CLS] token vector \cite{reimers-gurevych-2019-sentence,may-etal-2019-measuring}. In our experiments we select a linear kernel for the SVM and use PySVMRank,%
\footnote{PySVMRank, \url{https://github.com/ds4dm/PySVMRank}} 
a python API to the SVM$^{rank}$ library written in C.%
\footnote{SVM$^{rank}$, \url{www.cs.cornell.edu/people/tj/svm_light/svm_rank.html}} 
\section{Experiments and Discussion}
\label{sec:experiments}

We now present empirical experiments with the approaches from Section~\ref{sec:approach}. The goal is to evaluate how hard it is to compare and rank the claim revisions in our corpus from Section~\ref{sec:data} by quality.

\subsection{Experimental Setup}

We carry out experiments in two settings. The first considers creating {\em random splits} over revision histories, ensuring that all versions of the same claim are in a single split in order to avoid data leakage. We assign 80\% of the revision histories to the training set and the remaining 20\% to the test set. A drawback of this setup is that it is not clear how well models generalize to unseen debate categories. In the second setting, we therefore evaluate the methods also in a {\em cross-category} setup using a leave-one-category-out paradigm, which ensures that all claims from the same debate category are confined to a single split. We split the data in this way to evaluate if our models learn independent features that are applicable across the diverse set of categories. To assess the effect of adding augmented data, we evaluate all models on both ClaimRev\textsubscript{{\tiny {BASE}}} and ClaimRev\textsubscript{{\tiny {EXT}}}.

\begin{table*}[t]
\small
\centering
\setlength{\tabcolsep}{5pt}
\begin{tabular}{lrrrrrrrrrr}
\toprule
				& \multicolumn{5}{c}{\bf Random-Split}	& \multicolumn{5}{c}{\bf Cross-Category}	\\
				\cmidrule(l@{2pt}r@{2pt}){2-6}			\cmidrule(l@{2pt}r@{2pt}){7-11}				
\bf Model			& $r$	& $\rho$	& \bf Top-1	 & NDCG & MRR	& $r$	& $\rho$	& \bf Top-1	 & NDCG & MRR\\
\midrule
BTL + SBOW+L 			& 0.38	& 0.37	& 0.62		&  0.94 &0.79	&	0.36 	& 0.35	& 0.60	& 0.94 & 0.78	\\
BTL + BERT			& 0.60	& 0.59	&  0.74	& 0.96 &	0.86	&	0.58	& 0.57	& 0.72	 & 0.96 & 0.85	\\
BTL + SBERT			& 0.63	& 0.62	& 0.77	& \bf 0.97&	0.87	& 0.62	& 0.61	& 0.75	 &  0.97 & 0.86 	\\
\midrule
SVMRank + SBOW+L		&  0.18	& 0.18	& 0.50	& 0.93 & 0.73		&  0.24	& 0.23	& 0.52& 0.93 & 0.75	\\
SVMRank + BERT CLS	& 0.50	& 0.49	& 0.67	& 0.95 & 0.84	& 0.51&  0.51& 0.67 & 0.96 & 0.84	\\
SVMRank + SBERT		& \bf 0.70 & \bf 0.70	& \bf 0.79	& \bf 0.97&	\bf 0.90 	& \bf 0.73	&  \bf 0.72	& 	\bf 0.80&  \bf0.98 &  \bf0.91	\\
\midrule
Random 	baseline		& 0.00	& 0.00  & 0.42 & 0.91 & 0.68 & 0.00 & 0.00 & 0.42 &0.91 &	0.67	\\			
\bottomrule
\end{tabular}
\caption{Claim quality ranking results: Pearson's $r$ and Spearman's $\rho$ correlation as well as top-1 accuracy for all tested approaches in the random-split and the cross-category setting on ClaimRev\textsubscript{{\tiny{EXT}}}.In all cases, SVMRank + SBERT is significantly better than all others at $p <$ 0.001 according to a two-sided Student's $t$-test.}
\label{tab:ranking}
\end{table*}

For quality {\em classification}, we report accuracy and the Matthews correlation coefficient \cite{matthews-1975-mcc}. We report the mean results over five runs in the random setting and the mean results across all test categories in the cross-category setting. To ensure balanced class labels, we create one false claim pair for each true claim pair by shuffling the order of the claims: $(v_1, v_2, true) \rightarrow (v_2, v_1, false)$, where the label denotes whether the second claim in the pair is of higher quality. We report results obtained by models trained on ClaimRev\textsubscript{{\tiny {BASE}}} and ClaimRev\textsubscript{{\tiny {EXT}}} as score pairs in Table \ref{tab:res_class}.

To measure {\em ranking} performance, we calculate Pearson's $r$ and Spearman's $\rho$ correlation, as well as NDCG and MRR. We also compute the Top-1 accuracy, i.e. the proportion of claim sets, where the latest version has been ranked best. We average the results on each claim set across the test set for each metric. Afterwards we average the results across five runs or across all categories, depending on the chosen setting.

\subsection{Claim Quality Classification}

The results in Table \ref{tab:res_class} show that a claim's {\em length} is a weak indicator of quality (up to 61.3 accuracy). An intuitive explanation is that, even though claims with more information may be better, it is also important to keep them readable and concise.

Despite {\em SBOW}'s good performance on predicting convincingness \cite{potash-etal-2017-length}, the claim quality in our corpus cannot be captured by a model of such simplicity (maximum accuracy of 65.4). We point out that adding other linguistic features (for example, part-of-speech tags or sentiment scores) may further improve SBOW. Exemplarily, we equip SBOW with length features and observe a significant improvement (up to 67.5). 
 
As for the transformer-based methods, we see that {\em BERT} and {\em SBERT} consistently outperform SBOW in all settings on both corpus versions, with SBERT's accuracy of up to 77.7 being best.\footnote{Additionally, we have experimented with an adversarial training algorithm, ELECTRA \cite{Clark2020ELECTRA:}, and obtained results slightly better than BERT, yet inferior to SBERT. We omit to report these results here, since they did not provide any further notable insights.}

A comparison of the performance of the methods depending on the corpus used for training in Table~\ref{tab:res_class} shows the effect of augmenting the original Kialo data. In most cases, the results obtained by models trained on ClaimRev\textsubscript{{\tiny {EXT}}} are comparable (slightly higher/lower) than results obtained by models trained on ClaimRev\textsubscript{{\tiny{BASE}}}. This means that adding relations between non-consecutive claim versions does not improve the reliability of methods. Given that the performance scores obtained on the ClaimRev\textsubscript{{\tiny {EXT}}} test set are evidently higher than on the ClaimRev\textsubscript{{\tiny{BASE}}} test set, we can conclude that the augmented cases are easier to classify and the cumulative difference in quality is more evident.

We can also see in Table~\ref{tab:res_class} that the trained models are able to generalize across categories; the accuracy and MCC scores in the random split and cross-category settings for each method are very similar, with only a slight drop in the cross-category setting. This indicates that the nature of the revisions is relatively consistent among all categories, yet reveals the existence of some category-dependent features. 

To find out whether BERT really captures the relative revision quality and not only lexical features present in the original claim, we introduced a \textit{Single claim} baseline, analogous to the \textit{hypothesis-only} baseline in natural language inference\cite{poliak-etal-2018-hypothesis}. It can be seen that the accuracy and MCC scores are low across all settings (maximum accuracy of 59.8), which indicates that BERT indeed captures relative revision quality mostly.

\subsection{Claim Quality Ranking}

Table \ref{tab:ranking} lists the results of our ranking experiments, which show patterns similar to the results achieved in the classification task.

We can observe similar patterns in both of the selected ranking approaches: SBERT consistently outperforms all other considered approaches across all settings (up to 0.73 and 0.72 in Pearson's $r$ and Spearman's $\rho$ accordingly). BERT and SBERT outperform SBOW, indicating that transformer-based methods are more capable of capturing the relative quality of revisions. While BTL + BERT obtains results comparable to BTL + SBERT, we find that using the CLS-vector as a sentence embedding representation leads to lower results. We point out, though, that using other sentence embeddings and/or pooling strategies (for example, averaged BERT embeddings) may further improve results. 

Similar to the results of the classification task, we observe only a slight performance drop in the cross-category setting when using BTL for ranking, yet an increase when using SVMRank, again emphasizing the topic-independent nature of claim quality in our corpus.

\begin{table}[t]
\small
\centering
\setlength{\tabcolsep}{3pt}
\renewcommand{\arraystretch}{0.95}
\begin{tabular}{ll@{$\!\!\!\!$}rr}
\toprule				
\bf Task &       \bf Label & \bf Accuracy 	& \bf Instances	\\
\midrule
Type 	& Claim Clarification	& 69.7		& 12 856				\\
	 		& Typo/Grammar Correction	& 83.6		& 12 125		\\	
	 		& Corrected/Added Links    & 89.3 & 3 660		\\
	 		& Changed Meaning of Claim    & 57.3		& 232	\\
			& Misc	& 67.2		& 2 130  \\
	 		& None	& 78.3				& 45 842  \\
\midrule
Distance 		& Revision distance 1 	&  76.2  & 42 341 \\
			& Revision distance 2	 & 79.6 & 17 478 \\
			& Revision distance 3	 & 80.6  & 8 023	\\
			& Revision distance 4	 & 81.0  & 3 979	\\
			& Revision distance 5  	& 79.5 & 2 103	\\
			& Revision distance 6+  	& 74.9 & 2 921 \\
\midrule
			& \bf All 	& 	\bf 77.7 & \bf 76 845 \\			
\bottomrule
\end{tabular}
\caption{Accuracy of the best model, SBERT, on each single revision type and distance in ClaimRev\textsubscript{{\tiny{EXT}}}, along with the number of instances per each case.}
\label{tab:binary_error}
\end{table}

\subsection{Error Analysis}

To further explore the capabilities and limitations of the best model, SBERT, we analyzed its performance on each revision type and distance. 

As the upper part of Table \ref{tab:binary_error} shows, SBERT is highly capable of assessing revisions related to the correction and addition of links and supporting information. This revision type also obtained  the highest correlations between quality dimensions and type of revision (see Table~\ref{tab:dimension-analysis}), which indicates that the patterns of changes performed within this type are more consistent.
In contrast, we observe that the model fails to address revisions related to the changed meaning of a claim. On the one hand, this may be due to the fact that such examples are underrepresented in the data. On the other hand, the consideration of such examples in the selected tasks is questionable, since changing the meaning of claim is usually considered as the creation of a \textit{new claim} and not a \textit{new version} of a claim. 

An insight from the lower part of Table~\ref{tab:binary_error} is that the accuracy of predictions increases from revision distance~1 to~4. We obtain better results when comparing non-consecutive claims than when comparing claim pairs with distance of 1. An intuitive explanation is that, since each single revision should ideally improve the quality of a claim, the more revisions a claim undergoes, the more evident the quality improvement should be. For distances $> 5$, the accuracy starts to decrease  again, but this may be due to the limited number of cases given.

\section{Conclusion and Future Work}
\label{sec:conclusion}

In this paper, we have proposed a new way of assessing quality in argumentation by considering different revisions of the same claim. This allows us to focus on characteristics of quality regardless of the discussed topics, aspects, and stances in argumentation. We provide a new corpus of web claims, which is the first large-scale corpus to target quality assessment and revision processes on a claim level. We have carried out initial experiments on this corpus using traditional and transformer-based models, yielding promising results but also pointing to limitations. In a detailed analysis we have studied different kinds of claim revisions and provided insights into the aspects of a claim that influence the users' perception of quality. Such insights could help improve writing support in educational settings, or identify the best claims for debating technologies and argument search.

We seek to encourage further research on how to help online debate platforms automate the process of quality control and design automatic quality assessment systems. Such systems can be used to indicate if the suggested revisions increase the quality of an argument or recommend the type of revision needed. We leave it for future work to investigate whether the learned concepts of quality are transferable to content from other collaborative online platforms (such as idebate.org or Wikipedia), or to data from other domains, such as student essays and forum discussions.

\section*{Acknowledgments}
We thank Andreas Breiter for feedback on early drafts, and the anonymous reviewers for their helpful comments. This work was partially funded by the Deutsche Forschungsgemeinschaft (DFG, German Research Foundation) under project number 374666841, SFB 1342.

\bibliography{claim-quality}
\bibliographystyle{acl_natbib}


\end{document}